\documentclass{article}

\usepackage{arxiv}

\usepackage[utf8]{inputenc}
\usepackage[T1]{fontenc}
\usepackage{hyperref}
\usepackage{url}
\usepackage{booktabs}
\usepackage{amsfonts}
\usepackage{nicefrac}
\usepackage{microtype}
\usepackage{lipsum}
\usepackage{graphicx}
\graphicspath{ {./images/} }
\usepackage[numbers]{natbib}
\usepackage{subcaption}
\usepackage{amsmath}
\usepackage{enumitem}
\usepackage{listings}

\title{Content-Aware Ad Banner Layout Generation with Two-Stage Chain-of-Thought in Vision Language Models}

\author{Kei Yoshitake \\
Institute of Science Tokyo\\
{\tt\small yoshitake.k.6312@m.isct.ac.jp}\\
\and
Kento Hosono\\
HAKUHODO Technologies Inc.\\
{\tt\small kento.hosono@hakuhodo-technologies.co.jp}\\
\and
Ken Kobayashi \\
Institute of Science Tokyo\\
{\tt\small kobayashi.k@iee.eng.isct.ac.jp}
\and
Kazuhide Nakata \\
Institute of Science Tokyo\\
{\tt\small nakata.k.ac@m.titech.ac.jp}
}

\begin{document}
\maketitle
\begin{abstract}
In this paper, we propose a method for generating layouts for image‐based advertisements by leveraging a Vision–Language Model (VLM).
Conventional advertisement layout techniques have predominantly relied on saliency mapping to detect salient regions within a background image, but such approaches often fail to fully account for the image’s detailed composition and semantic content.
To overcome this limitation, our method harnesses a VLM to recognize the products and other elements depicted in the background and to inform the placement of text and logos.
The proposed layout‐generation pipeline consists of two steps. In the first step, the VLM analyzes the image to identify object types and their spatial relationships, then produces a text‐based “placement plan” based on this analysis.
In the second step, that plan is rendered into the final layout by generating HTML‐format code.
We validated the effectiveness of our approach through evaluation experiments, conducting both quantitative and qualitative comparisons against existing methods.
The results demonstrate that by explicitly considering the background image’s content, our method produces noticeably higher‐quality advertisement layouts.

\end{abstract}

\section{Introduction}
In recent years, the expansion of the digital advertising market has led to a growing demand for the creation of banner advertisements displayed on the web.
When creating banner ads manually, designers must select elements from a large pool of assets and repeatedly refine the layout through trial and error.
This process is both time- and labor-intensive, posing a major bottleneck for the large-scale production of banner ads.
Against this backdrop, there has been increasing interest in research on automated ad layout generation in recent years \cite{baykasouglu2001simulated,el2004genetic,hsu2023posterlayout,lin2023layoutprompter,li2020attribute,liu2023visual,tang2024layoutnuwa,yang2024posterllava,zhou2022composition,seol2025posterllama}.

A layout refers to the size and arrangement of elements such as text and logos that make up an ad.
A visually well-organized layout can effectively capture viewers’ attention and convey information.
Consequently, generating effective layouts is important not only for ads but also in a wide range of fields, including poster design, websites, and application user interface design.

\begin{figure}
\centering
\begin{minipage}[ht]{0.45\linewidth}
    \centering
    \includegraphics[keepaspectratio,scale=0.15]{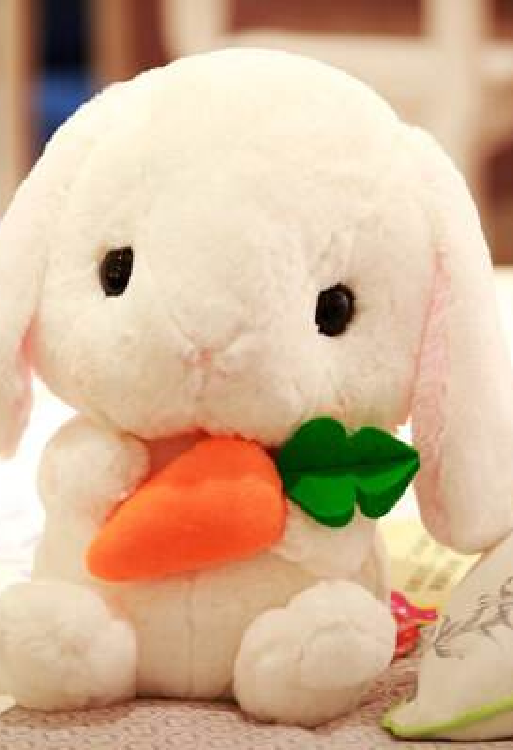}
    \subcaption{}
    \label{fig:a}
\end{minipage}
\begin{minipage}[ht]{0.45\linewidth}
    \centering
    \includegraphics[keepaspectratio,scale=0.15]{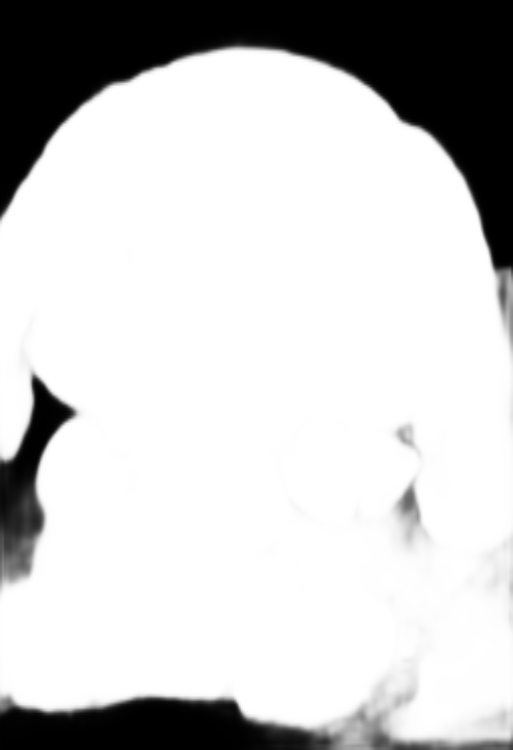}
    \subcaption{}
    \label{fig:b}
\end{minipage}
\caption{Example of salient region extraction using saliency mapping~\cite{hsu2023posterlayout}}
\label{fig:example_saliencymap}
\end{figure}

Among various use cases, banner ads require layout elements to be arranged so that they do not overlap with the products depicted in the background image.
Therefore, layout generation that takes into account the content of the background image is essential.
Previous studies have proposed layout generation methods based on saliency mapping \cite{hou2007saliency}, which identifies prominent regions within an image \cite{hsu2023posterlayout,lin2023layoutprompter,zhou2022composition}.
However, as shown in Figure~\ref{fig:example_saliencymap}, this approach has a limitation in that it cannot prioritize particularly important areas within the identified regions.
As a result, when the salient regions cover a wide area of the image, there may be insufficient space for layout elements, leading to overlaps with products or key objects.

To address this issue, we propose a layout generation method that leverages VLM\cite{tsimpoukelli2021multimodal,wang2025banneragency,liu2023visual}, which can process both visual and textual information in an integrated manner.
VLMs are capable of not only extracting visual features from an image but also deeply understanding the types of objects present, their spatial relationships, and the overall context as textual information.
By utilizing VLM, we aim to generate layouts that more effectively reflect the detailed composition and semantics of the background image.
However, directly generating layouts using VLM presents challenges: the model may interpret prompts in unintended ways, and controlling the final output can be difficult.
To overcome these issues, we adopt Chain-of-Thought (CoT) Prompting \cite{wei2022chain}, a technique designed to elicit more accurate outputs by guiding the model through an explicit reasoning process.
CoT prompting improves model comprehension and controllability in complex reasoning tasks by encouraging it to verbalize intermediate steps leading to a final decision.

Our proposed approach mitigates the black-box nature of VLM and allows for outputs that more accurately reflect the user’s intent. Specifically, we divide the layout generation task into two subtasks: 1) generating a text-based placement plan using the VLM, and 2) producing the final layout as HTML code based on the generated plan.
Through this two-step process, we enable more reliable and interpretable layout generation.

To validate the effectiveness of our proposed method, we conducted both quantitative and qualitative evaluations using a public dataset, comparing our approach against existing methods.

\section{Related Work}
Early layout generation methods formulated the placement of elements as an optimization problem and employed heuristic search techniques to solve it~\cite{baykasouglu2001simulated,el2004genetic}.
However, these approaches are limited in terms of the diversity and flexibility of the generated layouts.
With the advancement of deep learning, methods such as CGL-GAN~\cite{zhou2022composition} and DS-GAN~\cite{hsu2023posterlayout} have been proposed, which learn to generate layouts by leveraging visual information from images.
More recently, diffusion models~\cite{ho2020denoising,song2020denoising} have also been applied to layout generation~\cite{zheng2023layoutdiffusion,chai2023layoutdm,inoue2023layoutdm}.
These probabilistic generative models, which sample from learned data distributions, are capable of producing diverse layouts. 
However, they require substantial computational resources for both training and inference.
In addition, methods such as LayoutNUWA~\cite{tang2024layoutnuwa}, LayoutPrompter~\cite{lin2023layoutprompter}, and PosterLlama~\cite{seol2025posterllama} have demonstrated that high-quality layouts can be generated using large language models (LLMs).
These methods typically represent layouts in HTML format, thereby leveraging the code generation capabilities of LLMs.

In layout generation tasks involving background images, such as ads or posters, it is crucial to consider the semantic content of the image.
Approaches like PosterLayout~\cite{hsu2023posterlayout}, LayoutPrompter~\cite{lin2023layoutprompter}, and CGL-GAN~\cite{zhou2022composition} utilize saliency mapping~\cite{hou2007saliency} to extract visually prominent regions and design layouts that avoid them.
However, saliency maps do not fully capture the detailed semantic structure of the image, making it difficult to distinguish between regions that must be avoided (e.g., a character’s face) and those where overlap is acceptable.
For example, as shown in Fig.~\ref{fig:example_saliencymap}, bright regions in the saliency map are considered salient, but they do not differentiate between semantically important areas (e.g., a rabbit's face or a carrot) and less important ones.
As a result, when saliency maps highlight broad regions, it may be difficult to secure enough space for layout elements, leading to a significant decline in layout quality.

Recently, numerous vision-language models VLMs have been proposed~\cite{liu2023visual,achiam2023gpt,radford2021learning}, and their application to layout generation is gaining attention.
PosterLLaVa~\cite{yang2024posterllava} fine-tunes the VLM LLaVa~\cite{liu2023visual} and demonstrates the effectiveness of VLMs for layout generation that takes image content into account.

In this paper, instead of relying on saliency mapping, we propose a layout generation method that incorporates image understanding through VLMs.
Importantly, our method does not require any additional training or fine-tuning, aiming to enable general-purpose and lightweight layout generation.

\section{Methodology}
\subsection{Problem Formulation and Layout Representation}
Given an image, we define the horizontal direction as the $x$-axis and the vertical direction as the $y$-axis.
Each layout element $i$ is represented by its position $(x_i, y_i)$, width $w_i$, height $h_i$, and category $c_i$.
Denoting the element as $e_i = ((x_i, y_i, w_i, h_i), c_i)$, the entire layout can be defined as $L = {e_1, e_2, \ldots, e_N}$,
where $N$ is the number of elements included in the layout.
To leverage the code generation capabilities of large language models (LLMs), the layout is expressed in HTML format~\cite{lin2023layoutprompter,seol2025posterllama,tang2024layoutnuwa}.
Specifically, each $e_i$ is represented using an HTML \texttt{<div>} tag as follows:

\begin{align*}
\texttt{<div\ class="}\mathit{c_i}\texttt{" style="left:}\mathit{x_i}\texttt{px; top:}\mathit{y_i}\texttt{px; width:}\mathit{w_i}\texttt{px; height:}\mathit{h_i}\texttt{px"></div>}
\end{align*}

\subsection{Overview}
Our method consists of two steps:
1) generating a text-based placement plan using a VLM, and
2) generating an HTML-based layout based on the placement plan.
Here, the placement plan refers to a textual instruction describing where and how each element should be placed on the background image, as well as considerations to be taken into account.
By dividing layout generation into these two steps—planning and placement—we aim to make the model explicitly articulate its reasoning process, thereby enabling more appropriate and coherent outputs.
This approach is inspired by Chain-of-Thought (CoT) Prompting~\cite{wei2022chain}, which encourages the model to explicitly outline its reasoning steps, leading to improved performance on complex tasks.
In the following sections, we describe our method in detail.

\subsection{Placement Plan Generation}\label{layout-plan_generation}
To generate a placement plan, the VLM is given a background image, an instruction, and constraints specifying the categories and number of elements to be placed.
The model then outputs a placement plan in natural language format.
Few-shot prompting~\cite{brown2020language} is employed to guide the model toward the desired output by presenting several input-output pairs as exemplars.
This approach stabilizes the model’s responses and ensures that the output follows a consistent and interpretable structure.

The structure of the input and output examples is illustrated in Table~\ref{tab:planning prompt example}.
Each input example consists of a background image (\texttt{<image>}), a shared instruction for planning, a shared constraint encouraging attention to the image content, and an element type constraint that varies across examples.
The output is a text-based placement plan, manually created based on the actual layout.
By providing these elements—image, instruction, constraints, and element types—the prompt is designed to elicit placement plans that reflect the content and structure of the image.
In the experiments, the categories and numbers of elements were randomly sampled from the training data of the dataset.
During testing, the number of elements can be specified arbitrarily.

Each placement plan describes the preferred position of each element, taking into account factors such as overlap and visual balance, and includes placement considerations as needed.
This allows for the automatic generation of layout policies that reflect the semantic structure of the image through the use of a VLM.
A concrete example of a prompt used in placement plan generation is shown in Table~\ref{tab:planning prompt example}.

\begin{table}[htbp]
    \centering
    \caption{Example Prompt for Placement Plan Generation}
    \label{tab:planning prompt example}
    \begin{tabular}{p{16cm}}
        \hline
        Prompt \\ \hline
        \textbf{\textit{Instruction }}\\
        Please tell me the requirements where to place the ad elements. \\
        \textbf{\textit{Constraint}} \\
        - Please aware the contents of this image. \\
        \\
        \textbf{Example 1}\\
        \texttt{<image>} \\
        \textbf{\textit{Element Type Constraint}}: text 0 \textbar\ text 1 \textbar\ underlay 2 \\
        
        \\
        \textbf{Example Output 1} \\
        \textbf{\textit{Placement Plan}}: \\
        - Text 0 : Bottom center to ensure it dose not overlap above the waist of man and woman. \\
        - Text 1 : Under Text 0, aligned with Text 0. \\
        - Underlay 2 : Behind Text 0 and Text 1 to avoid overlapping humans as possible. \\
        \dots\dots\\
        \\
        \textbf{Test Sample}\\
        \texttt{<image>} \\
        \textbf{\textit{Element Type Constraint}}: text 0 \textbar\ text 1 \textbar\ underlay 2 \\ 
        \\ \hline
    \end{tabular}
\end{table}

\subsection{Layout Generation Based on Placement Plan}\label{layout_generation_with_plan}
In this step, a concrete layout is generated based on the placement plan produced in the first step.
The input to the VLM consists of the background image, the placement plan, the element type constraints, and the same background and constraint information used in Section~\ref{layout-plan_generation}.
Given this input, the VLM generates the layout as HTML code specifying the element positions.
As in the first step, a few input-output examples are included in the prompt to guide the generation process.

Table~\ref{tab:prompt example} shows the format of the prompt used in this step.
Each input example includes the background image (\texttt{<image>}), shared instructions for layout generation (Instruction, Task Description, Canvas Size), an element type constraint (Element Type Constraint), and the placement plan (Placement Plan) used in Section~\ref{layout-plan_generation}.
The corresponding output is the layout expressed in HTML format, as shown in Example Output 1 of Table~\ref{tab:prompt example}.
In this paper, the VLM is prompted with pairs of manually written placement plans and their corresponding HTML layouts, enabling the model to generate layouts automatically in response to the given image.

Finally, the HTML-based layout generated by the VLM is rendered on top of the background image to visualize the resulting ad layouts.
Table~\ref{tab:prompt example} provides a concrete example of the prompt used for layout generation based on the plan.

\begin{table}[htbp]
    \centering
    \caption{Example Prompt for Layout Generation Based on the Plan}
    \label{tab:prompt example}
    \begin{tabular}{p{16cm}}
        \hline
        Prompt \\ \hline
        \textbf{\textit{Instruction}}: Please generate a layout based on the given information. You need to ensure that the generated layout looks realistic, with elements well aligned and avoiding unnecessary overlap. \\
        \textbf{\textit{Task Description}}: content-aware layout generation \\
        Please place the following elements to avoid salient content, and underlay must be the background of text or logo. \\
        \textbf{\textit{Canvas Size}}: canvas width is 102px, canvas height is 150px\\
        \\
        \textbf{Example 1}\\
        \texttt{<image>} \\
        \textbf{\textit{Element Type Constraint}}: text 0 \textbar\ text 1 \textbar\ underlay 2 \\
        \textbf{\textit{Placement Plan}}: \\
        - Text 0 : Bottom center to ensure it dose not overlap above the waist of man and woman. \\
        - Text 1 : Under Text 0, aligned with Text 0. \\
        - Underlay 2 : Behind Text 0 and Text 1 to avoid overlapping humans as possible. \\
        \\
        \textbf{Example Output 1}
        \begin{verbatim}
<html>
<body>
<div  class="canvas" style="left:0px; top:0px; width:102px; height:150px"></div>
<div  class="text" style="left:2px; top:113px; width:95px; height:10px"></div>
<div  class="text" style="left:2px; top:124px; width:95px; height:9px"></div>
<div  class="underlay" style="left:0px; top:111px; width:102px; height:24px"></div>
</body>
</html>
        \end{verbatim}
        \vspace{-1em}
        \dots\dots\\
        \\
        \textbf{Test Sample}\\
        \texttt{<image>} \\
        \textbf{\textit{Element Type Constraint}}: text 0 \textbar\ text 1 \textbar\ underlay 2 \\
        \textbf{\textit{Placement Plan}}: \\
        - Text 0: Bottom center, below the plush toy. \\
        - Text 1: Below text 0, aligned to the left. \\
        - Underlay 2: Behind text 2 for contrast. \\ 
        \\ \hline
    \end{tabular}
\end{table}

\subsection{Comparison with One-Step CoT Prompting}
As described in this section, our method adopts a two-step prompting strategy that separates layout planning from element placement.  
To evaluate the effectiveness of this approach, a baseline using the conventional one-step Chain-of-Thought (CoT) prompting method is introduced for comparison.  
A quantitative and qualitative evaluation of the two approaches is presented in Section~\ref{Experiments}.  
In the one-step method, both the placement plan and the final placement are generated simultaneously within a single output.

\section{Experiments}\label{Experiments}

\subsection{Setups}
\begin{description}
    \item[Dataset]
    This paper uses the publicly available ad dataset PKU-PosterLayout~\cite{hsu2023posterlayout}.
    The dataset classifies layout elements into three categories: logo, text, and underlay. Each sample consists of an ad image along with category labels and bounding box coordinates for each element.
    The training set contains 9,974 image–layout pairs, while the test set includes 905 images.
    In the experiments, a subset of the training data is used for exemplars, and layout generation is performed on 100 test samples.

    \item[Evaluation Metrics] 
    We adopt eight evaluation metrics commonly used in prior studies~\cite{hsu2023posterlayout,lin2023layoutprompter,yang2024posterllava}: Validity, Overlap, Alignment, Underlay Loose, Underlay Strict, Utility, Occlusion, and Unreadability.

  \begin{itemize}
    \item \textbf{Validity:} Measures the proportion of valid elements, excluding those that are too small or fall outside the image boundaries.
    \item \textbf{Overlap:} Applies penalties for overlaps between elements, excluding underlays.
    \item \textbf{Alignment:} Penalizes misalignment between elements in left, center, or vertical directions.
    \item \textbf{Underlay Loose:} Evaluates the inclusion of other elements within the underlay using the maximum intersection-over-union (IoU).
    \item \textbf{Underlay Strict:} Evaluates the stricter condition of full containment of other elements by the underlay.
    \item \textbf{Utility:} Measures the efficiency of space usage by calculating the proportion of elements placed outside salient regions.
    \item \textbf{Occlusion:} Penalizes elements that overlap with salient regions of the saliency map.
    \item \textbf{Unreadability:} Assesses potential degradation in text readability by measuring gradient variation in the background outside the underlay.
  \end{itemize}

    \item[Baselines]  
    We compared our proposed method with three existing approaches.  
    The first is DS-GAN~\cite{hsu2023posterlayout}, which employs a CNN encoder and a CNN-LSTM decoder.  
    The second is LayoutPrompter~\cite{lin2023layoutprompter}, which uses a pre-trained LLM to generate layouts in HTML format via few-shot prompting.  
    Both DS-GAN and LayoutPrompter utilize saliency maps to incorporate image content into layout generation.  
    The third method is PosterLLaVa~\cite{yang2024posterllava}, which fine-tunes a VLM to perform image understanding and layout generation in an end-to-end manner.  
    We conducted a quantitative comparison between our method and these baselines.

    \item[Implementation Details]  
    We used GPT-4o~\cite{achiam2023gpt} by OpenAI as the vision-language model, specifically the model version \texttt{gpt-4o-2024-08-06}.  
    In the original LayoutPrompter paper, GPT-3 text-davinci-003 was used as the LLM. However, since that model is no longer available, we used GPT-4o as a substitute for reproduction and comparison.  
    Inference was performed with the OpenAI API using its default parameters, where temperature was set to 0.7 and top\_p was set to 1.

\end{description}

\subsection{Ablation Study}

To evaluate the effectiveness of our method, we conducted a comparative experiment by modifying certain components of the method.  
Layouts were generated under the following seven conditions, and the results were compared.

\begin{enumerate}[label=(\alph*)]
    \item Baseline: One-step generation (0-shot, no CoT, no planning)  
    \item Comparison 1: One-step generation (0-shot, with CoT)  
    \item Comparison 2: Two-step generation (0-shot, with CoT)  
    \item Comparison 3: One-step generation (5-shot, with CoT)  
    \item Comparison 4: One-step generation (10-shot, with CoT)  
    \item Comparison 5: Two-step generation (5-shot, with CoT)  
    \item Comparison 6: Two-step generation (10-shot, with CoT)  
\end{enumerate}

We refer to the number of input-output examples in the prompt as N-shot (e.g., 0-shot, 5-shot, 10-shot).

\subsection{Results}
\subsubsection{Quantitative Evaluation} \label{subsubsec: quantitative_evaluation}
Tables~\ref{tab:result_evaluation} and~\ref{tab:result_comparative approach} present the results of the quantitative evaluation.  
The best scores are shown in bold, and the second-best scores are underlined.  
The results for previous methods shown in Table~\ref{tab:result_evaluation} are taken from the respective original papers.  
“LayoutPrompter (4o)” in the table refers to the reproduction of LayoutPrompter using GPT-4o.

In Table~\ref{tab:result_evaluation}, our method achieves comparable or superior performance in most metrics, although its scores for \textbf{Alignment} and \textbf{Utility} are slightly lower than those of the baselines.  
However, lower values in Alignment and Utility do not necessarily indicate poor visual design. Upon visual inspection, the layouts generated by our method were found to maintain sufficient visual balance.

Compared to LayoutPrompter, our method performs worse on Utility and Occlusion, both of which reflect the model's ability to consider image content.  
This is likely because these metrics are computed using saliency maps, which are explicitly incorporated in LayoutPrompter but not used in our method.

\begin{table}[ht]
  \centering
  \caption{Quantitative Comparison with baselines on the PosterLayout dataset.}
  \label{tab:result_evaluation}
  \resizebox{\linewidth}{!}{  
    \begin{tabular}{lcccccccc}
      \toprule
      & Val $\uparrow$ & Ove $\downarrow$ & Ali $\downarrow$ & Und$_l$ $\uparrow$ & Und$_s$ $\uparrow$ & Uti $\uparrow$ & Occ $\downarrow$ & Rea $\downarrow$ \\
      \midrule
      DS-GAN\cite{hsu2023posterlayout} & 0.8788 & 0.0220 & 0.0046 & 0.8315 & 0.4320 & 0.2541 & 0.2088 & 0.1874 \\
      LayoutPrompter\cite{lin2023layoutprompter} & 0.9992 & 0.0036 & \underline{0.0036} & 0.8986 & 0.8802 & \underline{0.2597} & \textbf{0.0992} & 0.1723 \\
      LayoutPrompter(4o)
      \cite{lin2023layoutprompter} & \textbf{1.0000} & 0.0032 & 0.0076 & 0.7899 & 0.7485 & 0.2337 & 0.3895 & 0.1749 \\
      PosterLLaVa\cite{yang2024posterllava} & \textbf{1.0000} & \textbf{7.7e-5} & \textbf{0.0002} & \textbf{1.0000} & \textbf{1.0000} & \textbf{0.2628} & 0.1649 & \underline{0.1142} \\
      Ours & \textbf{1.0000} & \underline{4.1e-9} & 0.0142 & \underline{0.9882} & \underline{0.9642} & 0.1696 & \underline{0.1619} & \textbf{0.0932} \\
      \bottomrule
    \end{tabular}
  }
\end{table}

Focusing on Table~\ref{tab:result_comparative approach}, the values of Utility and Occlusion tend to be higher in conditions (a), (b), and (c), which are generated without few-shot examples.  
High values in both metrics suggest that the model failed to sufficiently consider the image content.  
These conditions also exhibit low Underlay scores, indicating that the underlay elements were not appropriately placed behind other elements.

Next, Comparisons 3-6 are examined.  
There is no substantial difference in the metrics related to element relationships—Validity, Overlap, Alignment, and Underlay—between one-step and two-step methods.  
However, two-step prompting results in improved Occlusion scores, suggesting that it enables more accurate layout generation that better reflects image content.  
Changing the number of examples (5-shot vs. 10-shot) did not lead to significant differences in the quantitative results.

\begin{table}[ht]
  \centering
  \caption{Quantitative Comparison with our method on the PosterLayout dataset}
  \label{tab:result_comparative approach}
  \resizebox{\linewidth}{!}{
    \begin{tabular}{lcccccccc}
      \toprule
      & Val $\uparrow$ & Ove $\downarrow$ & Ali $\downarrow$ & Und$_l$ $\uparrow$ & Und$_s$ $\uparrow$ & Uti $\uparrow$ & Occ $\downarrow$ & Rea $\downarrow$ \\
      \midrule
      Baseline (0-shot, no CoT) & 1.0000 & 5.7e-10 & 0.0002 & 0.4694 & 0.4694 & 0.5437 & 0.3625 & 0.1231 \\
      Comparison 1 (1-step, 0-shot) & 1.0000 & 0.0150 & 0.0144 & 0.5096 & 0.4899 & 0.4401 & 0.3144 & 0.1399 \\
      Comparison 2 (2-step, 0-shot) & 1.0000 & 0.0071 & 0.0443 & 0.8804 & 0.8650 & 0.3203 & 0.2569 & 0.1236 \\ \midrule
      Comparison 3 (1-step, 5-shot) & \textbf{1.0000} & 0.0029 & \underline{0.0187} & 0.9974 & 0.9735 & \underline{0.1838} & 0.2502 & 0.1394 \\
      Comparison 4 (1-step, 10-shot) & \textbf{1.0000} & \textbf{0.0000} & 0.0251 & 0.9816 & \underline{0.9742} & \textbf{0.2501} & 0.2526 & \textbf{0.0902} \\
      Comparison 5 (2-step, 5-shot) & \textbf{1.0000} & \underline{4.1e-9} & \textbf{0.0142} & \underline{0.9882} & 0.9642 & 0.1696 & \textbf{0.1619} & \underline{0.0932} \\
      Comparison 6 (2-step, 10-shot) & 0.9984 & 0.0043 & 0.0330 & \textbf{0.9994} & \textbf{0.9943} & 0.1642 & \underline{0.1675} & 0.1082 \\
      \bottomrule
    \end{tabular}
  }
\end{table}

\subsubsection{VLM Evaluation}
As noted in Section~\ref{subsubsec: quantitative_evaluation}, widely used rule-based evaluation metrics do not always align with human perception.  
To address this issue, VLM-based evaluation has recently gained attention, with several studies reporting higher correlation with human judgment compared to rule-based metrics~\cite{haraguchi2024can, wang2025banneragency}.  
In this paper, we applied the evaluation method proposed by Haraguchi et al.~\cite{haraguchi2024can} to layouts generated by our method and conducted two types of analysis using VLM: absolute evaluation and relative (pairwise) evaluation.

Following prior work, we used GPT-4o as the VLM.  
Unlike previous studies that evaluated real ad images, our method generates layouts where each element is represented as a rectangular box.  
To account for this difference, the prompts given to GPT-4o explicitly state that the layout is represented in box format.  
An example of the prompts used is provided in the Appendix.

In the absolute evaluation, each layout was individually scored by GPT-4o on a scale from 1 to 10 based on three criteria: \textbf{Alignment}, \textbf{Overlap}, and \textbf{White Space}.  
GPT-4o was also instructed to provide a justification for each score, which was used to verify the validity of the evaluations.  
Here, Alignment measures visual harmony and stability based on horizontal and vertical alignment of elements; Overlap evaluates the degree to which unnecessary overlaps impair readability; and White Space assesses whether adequate margins are maintained.

\begin{figure}
    \centering
    \includegraphics[width=0.6\linewidth]{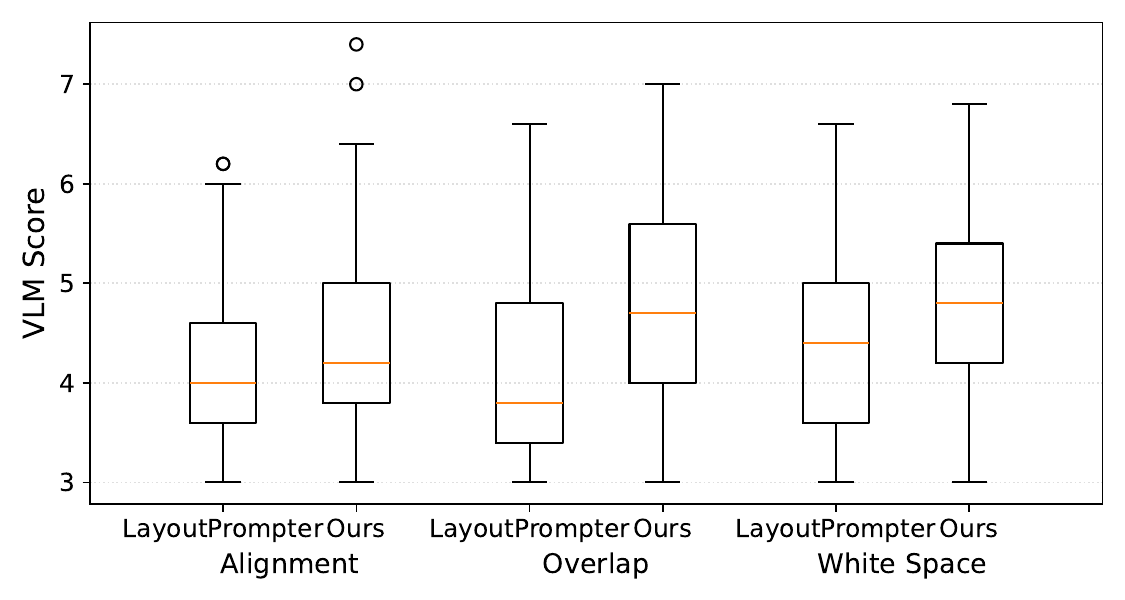}
    \caption{Box plots of GPT-4o evaluation scores}
    \label{fig:boxplot_vlm_eval}
\end{figure}

Figure~\ref{fig:boxplot_vlm_eval} shows box plots of the score distributions for each metric.  
Across all metrics, our method shows higher medians and broader upper quartile ranges compared to LayoutPrompter.  
In particular, the distributions for Overlap and White Space are more skewed toward higher scores, suggesting that our method tends to produce higher-quality layouts in these aspects.

\begin{figure}
    \centering
    \includegraphics[width=0.95\linewidth]{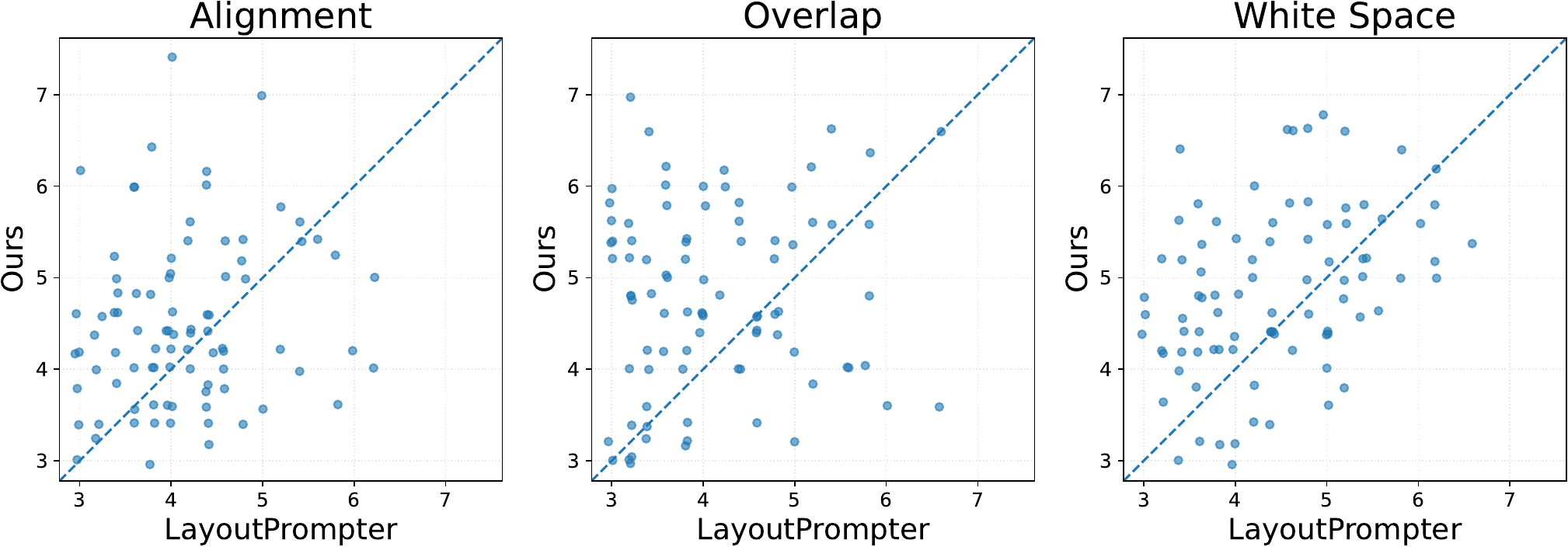}
    \caption{Scatter plots of GPT-4o evaluation scores}
    \label{fig:scatter_vlm_eval}
\end{figure}

Figure~\ref{fig:scatter_vlm_eval} presents scatter plots comparing the absolute scores of layouts generated by our method and LayoutPrompter.  
Although the difference is not significant in Alignment, a large number of data points lie above the 45-degree diagonal line for Overlap and White Space, indicating that our method tends to receive higher scores.
These results suggest that the layouts generated by our method are more highly rated by GPT-4o.

In the relative evaluation, layout pairs generated by our method and LayoutPrompter were presented to GPT-4o.  
For each metric, GPT-4o was asked to judge which layout was better and to provide a justification for its decision.  
Table~\ref{tab:vlm_pairwise_evaliation} shows the results of this pairwise evaluation.  
In all metrics, our method was preferred more often than LayoutPrompter, with particularly notable differences in Overlap and White Space.

\begin{figure}[ht]
  \centering
  \begin{minipage}[c]{0.3\linewidth}
    \centering
    \includegraphics[width=\linewidth]{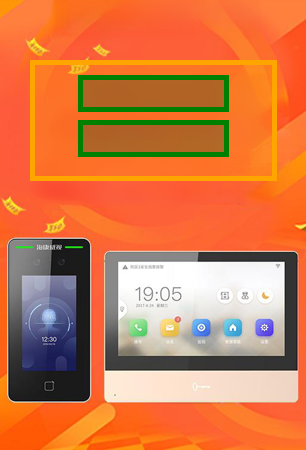}
  \end{minipage}
  \hfill
  \begin{minipage}[c]{0.65\linewidth}
    \small
    \textbf{Explanation:} \\
    The design shows a clear misuse of overlap and layering principles. Firstly, the text (green boxes) is not supported by a yellow underlay as intended, leading to poor contrast and readability issues. The absence of a yellow underlay beneath the text elements suggests a critical oversight in ensuring text visibility. ...\textless omitted \textgreater\\
    \textbf{Overlap score: 3.0}
  \end{minipage}
  \caption{Example of potentially inappropriate evaluation by GPT-4o}
  \label{fig:vlm_misjudge_example}
\end{figure}

These results indicate that our method tends to be rated more favorably than existing methods when evaluated by GPT-4o.  
However, some cases were observed where GPT-4o’s evaluation appeared to be inappropriate.  
Figure~\ref{fig:vlm_misjudge_example} illustrates one such case.  
Although the text element (green box) is properly overlaid on an underlay (yellow) for visual support, GPT-4o failed to recognize this and assigned a low score, citing insufficient contrast with the background.  
This result suggests that directly applying existing VLM-based evaluation methods~\cite{haraguchi2024can} to this task may not fully capture the visual context or layout structure.  
This limitation is particularly important in our setting, where layout elements are represented as rectangular boxes, whereas previous studies used actual ad images as input.

\begin{table}[ht]
  \centering
  \caption{GPT-4o pairwise preference ratios}
  \label{tab:vlm_pairwise_evaliation}
  \begin{tabular}{lcc}
    \toprule
    Metrics & LayoutPrompter(4o) & Ours \\
    \midrule
    Alignment     & 40.2 & 59.8 \\
    Overlap       & 27.6 & 72.4 \\
    White Space   & 33.3 & 66.7 \\
    \bottomrule
  \end{tabular}
\end{table}

\subsubsection{Qualitative Evaluation}

\begin{figure}[htbp]
    \centering
    \includegraphics[width=0.95\linewidth]{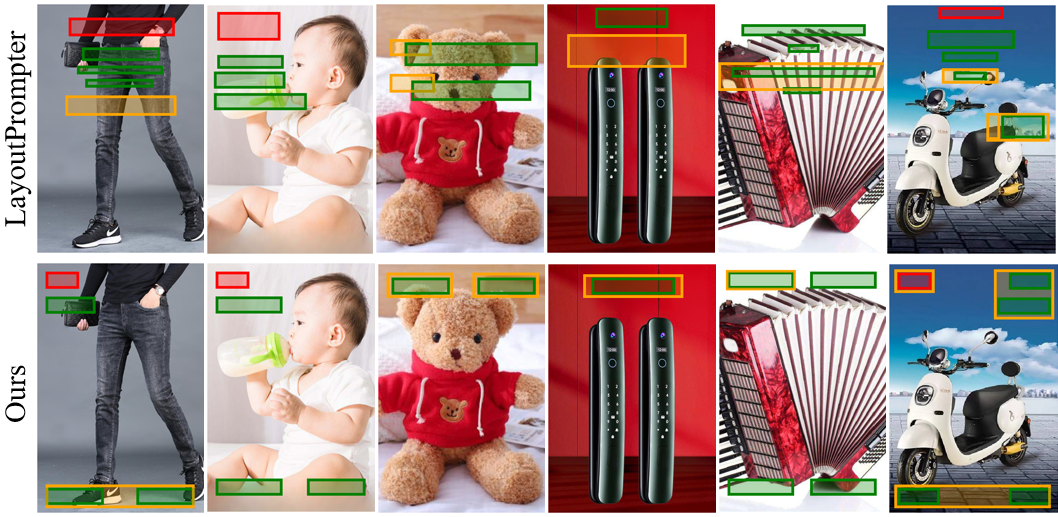}
    \caption{Qualitative results generated by LayoutPrompter and ours}
    \label{fig:layoutprompter_vs_ours}
\end{figure}

Figure~\ref{fig:layoutprompter_vs_ours} presents examples of layouts generated by LayoutPrompter and our proposed method.  
In these examples, LayoutPrompter sometimes places elements over important regions of the image, resulting in visually suboptimal designs.  Additionally, violations of placement constraints—such as underlay elements not being positioned beneath logos or text—are frequently observed in the outputs of LayoutPrompter.  
In contrast, our method successfully avoids salient regions and places elements in less important areas, such as the lower parts of the image.  
Constraint violations are rarely observed in the outputs of our method, and the percentage of violations, as evaluated by rule-based constraint checking, remains low (see Table~\ref{tab:constraint_violation}).  
These results indicate that our method, which leverages VLMs, is capable of generating layouts that are visually more appropriate.

\begin{table}[t]
  \centering
  \caption{Comparison of constraint violation rates (percentage of layouts with violations among all generated samples)}
  \label{tab:constraint_violation}
  \begin{tabular}{lcc}
    \toprule
     Method & Violation Rate (\%) \\
    \midrule
    LayoutPrompter & 21.88 \\
    Comparison 3 (1-step, 5-shot) & 3.00 \\
    Comparison 4 (1-step, 10-shot) & 0.00 \\
    Comparison 5 (2-step, 5-shot) & 1.00 \\
    Comparison 6 (2-step, 10-shot) & 0.00 \\
    \bottomrule
  \end{tabular}
\end{table}

\begin{figure}[ht]
    \centering
    \includegraphics[width=0.7\linewidth]{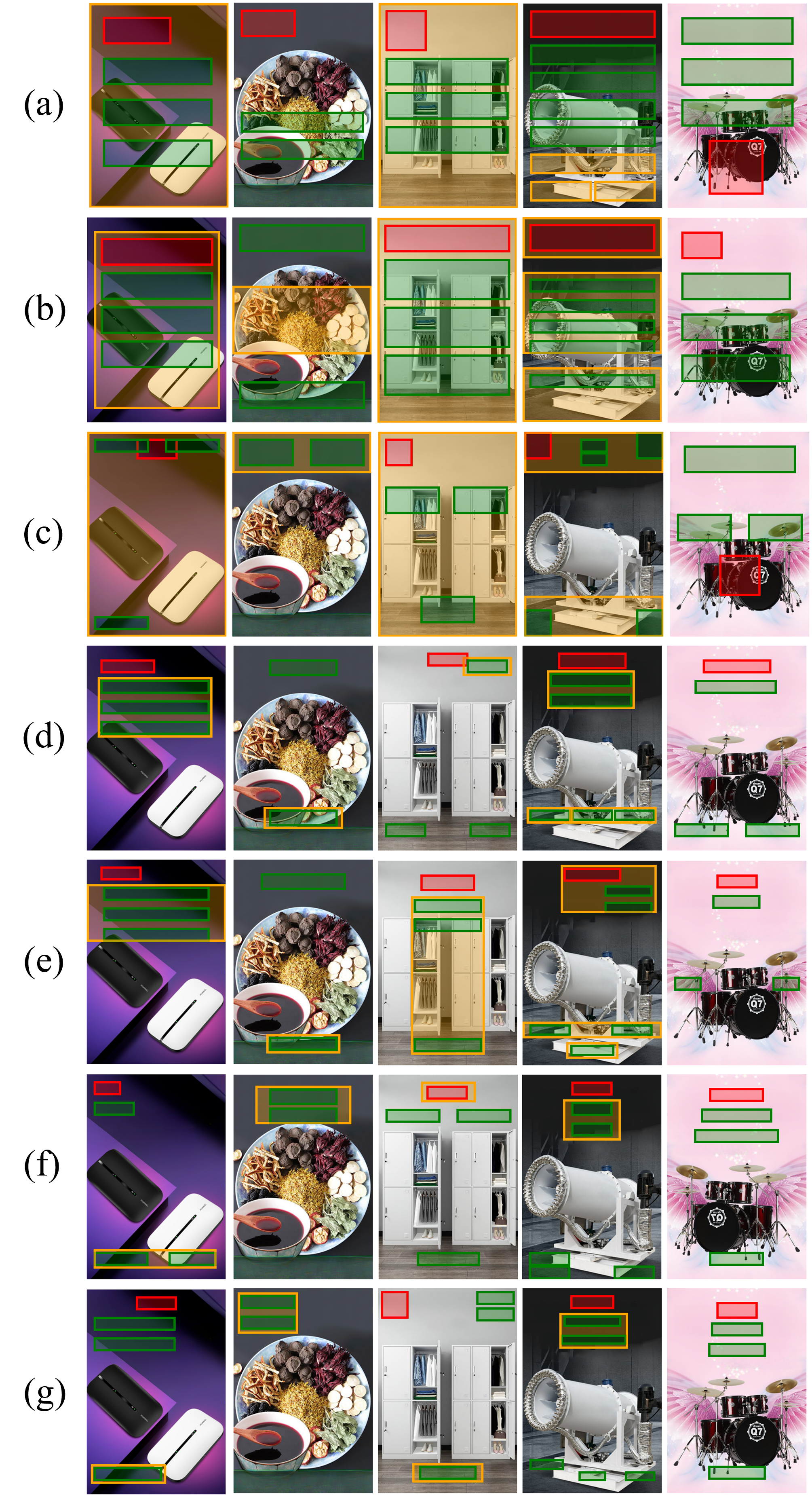}
    \caption{Qualitative results for the proposed method with varying settings}
    \label{fig:layout samples}
\end{figure}

Figure~\ref{fig:layout samples} shows layout examples generated under modified settings of the proposed method.  
When no examples are provided (0-shot), elements frequently overlap with important parts of the image, indicating that image content is not sufficiently taken into account.

\begin{figure}[ht]
    \centering
    \begin{minipage}[c]{0.3\linewidth}
      \centering
      \includegraphics[width=\linewidth]{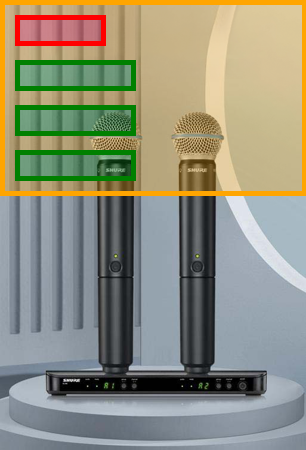}
    \end{minipage}
    \hfill
    \begin{minipage}[c]{0.65\linewidth}
      \begin{lstlisting}[basicstyle=\footnotesize\ttfamily, frame=single, breaklines=true]
Below is the HTML layout with elements placed to avoid overlapping the microphones and their base.
<html>
...
</html>
- Logo 0: Positioned at the top left to avoid the microphones.
- Text 1, 2, 3: Stacked vertically on the left side, avoiding the microphones.
- Underlay 4: Covers the area behind the text, not overlapping the microphones.
      \end{lstlisting}
    \end{minipage}
\caption{Output of the one-step method: layout results (top) and model output including plan (bottom)}
\label{fig:onestep_plan}
\end{figure}

Furthermore, comparing the one-step and two-step generation settings, the two-step method more effectively places elements in less salient areas, suggesting that it better incorporates image content into the layout.
To analyze this difference in more detail, we examined the model outputs for both the placement plan and the corresponding HTML layout.
In the one-step generation setting, as shown in Figure~\ref{fig:onestep_plan}, we frequently observed cases in which the HTML layout was generated first, followed by a textual placement plan.  
This output pattern suggests that the model may generate plausible HTML code upfront and then attempt to retroactively align the placement plan with the code.  
In fact, in the example shown in Figure~\ref{fig:onestep_plan}, text and underlay elements are placed without avoiding the microphone area, which is a visually important region.  
Such behavior—generating code first and then producing a plan to match it—may compromise both image awareness and constraint satisfaction, ultimately resulting in visually inappropriate layouts.

\begin{figure}[ht]
    \centering
    \begin{minipage}[c]{0.3\linewidth}
      \centering
      \includegraphics[width=\linewidth]{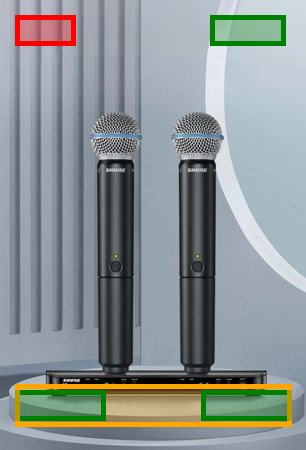}
    \end{minipage}
    \hfill
    \begin{minipage}[c]{0.65\linewidth}
      \begin{lstlisting}[basicstyle=\footnotesize\ttfamily, frame=single, breaklines=true]
Two microphones are placed on a platform in the center of the image. Avoid overlapping the microphones and the platform.
- Logo 0 : Top left, avoiding the microphones.
- Text 1 : Top right, avoiding the microphones.
- Text 2 : Bottom left, near the base, avoiding the platform.
- Text 3 : Bottom right, near the base, avoiding the platform.
- Underlay 4 : Behind text 2 and text 3, not to overlap the microphones or platform.
      \end{lstlisting}
    \end{minipage}
\caption{Output of the two-step method: layout results (top) and model-generated placement plan (bottom)}
\label{fig:twostep_plan}
\end{figure}

In contrast, the two-step generation approach explicitly separates the planning and layout phases: the model first generates a placement plan that considers the image content, and then produces the HTML layout based on that plan.  
This process ensures that the rationale for placement is established in advance, leading to more consistent code generation.  
As a result, the layout is more likely to avoid important objects in the image, producing visually appropriate designs with greater precision.  
As illustrated in Figure~\ref{fig:twostep_plan}, layouts generated using the two-step approach more reliably place elements outside of visually salient regions compared to the one-step approach.

\section{Conclusion}

In this paper, we proposed a method for generating visually coherent banner ad layouts using a VLM, with the goal of automating the layout design process while taking the content of the background image into account.  
By prompting the VLM to generate placement plans, our method enables layout generation that reflects the structural content of the image—something that conventional saliency-based methods cannot fully capture.
The proposed method does not rely on large-scale fine-tuning.  
Instead, it combines a two-step generation process consisting of text-based placement planning and HTML-based code generation, making it effective even in low-data scenarios.  
We also validated the effectiveness of our method through comparative experiments with prior methods and ablated versions of our own approach.
In quantitative evaluations, although our method showed slightly lower performance in metrics related to alignment and whitespace usage, it achieved competitive or superior results in other metrics.  
Qualitative evaluations further confirmed that the method is capable of generating layouts that appropriately consider the visual content of the image.
Additionally, analysis using VLM-based evaluation metrics showed that our method tends to receive higher scores from GPT-4o compared to existing approaches.  
However, some cases were observed in which the evaluation did not align with the actual layout quality, indicating that current VLM-based evaluation methods still have room for improvement.

{
    \small
    \bibliographystyle{ieeenat_fullname}
    \bibliography{main}
}

\end{document}